# TOPSIS-RAD: RANKING ACCORDING TO DESIRES

A PREPRINT

**Leonardo Fernandes Costa**
Universidade Federal Fluminense
Rua Passos da Pátria, 156
Niterói, RJ , Brazil, CEP 24.220-045
Leonardo Sistemas Consultoria LTDA
Av. Albert0 Lamego, 405
Campos dos Goytacazes, RJ, 28016-811, Brazil
leo_costa@id.uff.br

**Helder Gomes Costa**
Universidade Federal Fluminense
Rua Passos da Pátria, 156
Niterói, RJ , Brazil, CEP 24.220-045
heldergc@id.uff.br [Corresponding author]

**Diogo Lima**
Universidade Federal Fluminense
Rua Passos da Pátria, 156
Niterói, RJ , Brazil, CEP 24.220-045
diogofls@id.uff.br

**Brunno Rodrigues**
Rua Passos da Pátria, 156
Niterói, RJ , Brazil, CEP 24.220-045
brunnoesr@id.uff.br

**ABSTRACT**

Traditional TOPSIS derives its reference points—the Positive Ideal Solution ($PIS$) and Negative Ideal Solution ($NIS$)—from the observed alternative set, making rankings susceptible to misalignment with decision-maker (DM) requirements, sensitivity to outlier performances, and rank reversal. This paper proposes TOPSIS-RAD, which addresses these issues by incorporating two arrays of DM-defined reference levels. Vetoed Performance Levels ($VPL$) exclude non-viable alternatives before normalisation, preventing them from distorting the ranking frontiers. Desired Performance Levels ($DPL$) cap performances at the DM's desired level before normalisation, anchoring the $PIS$ in explicit aspirations rather than dataset extremes. Three toy examples demonstrate each mechanism: $VPL$ reshapes normalisation boundaries by removing a non-viable alternative; fixed $DPL$ frontiers stabilise rankings by limiting the influence of performances well above the desired level. The method preserves the familiar distance-based structure of TOPSIS while grounding the ranking in stable, DM-specified boundaries. Limitations and future research directions are also discussed.

## 1 Introduction

The Technique for Order of Preference by Similarity to Ideal Solution (TOPSIS), introduced by Hwang and Yoon [1981], remains one of the best-known methods in Multi-Criteria Decision Aid/Making (MCDA/M). Its logic is straightforward: alternatives are ranked according to their proximity to two reference points, the Positive Ideal Solution ($PIS$), representing the best attainable performance across all criteria, and the Negative Ideal Solution ($NIS$), representing the worst performance.

Under this rationale, the preferred alternative should be as close as possible to the $PIS$ and as far as possible from the $NIS$. That dual-distance rule made TOPSIS influential in the MCDA/M literature, but it also exposed the method to recurring criticisms. As discussed in section 3, the main concerns can be summarized as follows:

(i) Misalignment with decision-maker (DM) preferences - The $PIS$ and $NIS$ are artificially constructed and may not reflect the decision-maker's actual needs or aspirations. Consequently, these idealized frontiers can diverge from practical or context-specific requirements [Behzadian et al., 2012].

(ii) Sensitivity to outliers - The presence of outliers in the set of alternatives can distort the calculation of $PIS$ and $NIS$, introduce noise, and potentially lead to anomalies such as rank reversal [Anes and Abreu, 2025].

(iii) Rank reversal phenomenon - Because the $PIS$ and $NIS$ are derived from the performances observed in the initial set of alternatives, adding or removing an alternative, which performance is part o PIS or NIS, can alter the relative distances of the remaining options. This dependency may change the ranking order and undermine the stability of the method [Kong, 2011, Chen et al., 2011, García-Cascales and Lamata, 2012a, de Farias Aires and Ferreira, 2019].

This paper addresses these limitations by introducing a TOPSIS variant, referred to as TOPSIS-RAD. The key move is simple: before the final distance-based ranking stage, the decision maker specifies two reference levels that reshape the decision problem itself. More specifically, the proposed method is intended to:

(i) Incorporate DM preferences - Ensure that the ranking process reflects the DM's actual needs and aspirations rather than relying on artificially constructed ideal points.

(ii) Mitigate the impact of outliers - Reduce sensitivity to extreme values that can distort rankings and compromise the stability of the decision-making process.

(iii) Enhance robustness and practical relevance - Provide a ranking framework grounded on explicit and stable minimum acceptable and desired performance levels, instead of based on PIS and NIS.

To do so, TOPSIS-RAD adds two elements to the standard TOPSIS workflow:

- Desired Performance Levels ($DPL$) - An array representing the DM's preferred performance goals, used to cap performances above the desired level before normalization and ranking.
- Vetoed Performance Levels ($VPL$) - An array defining minimum acceptable performance thresholds, used to exclude non-viable alternatives before ranking.

Accordingly, TOPSIS-RAD extends TOPSIS by changing how the ranking problem is constructed before the final distance-based scoring stage.

Figure 1 presents an overview of how the article is structured.

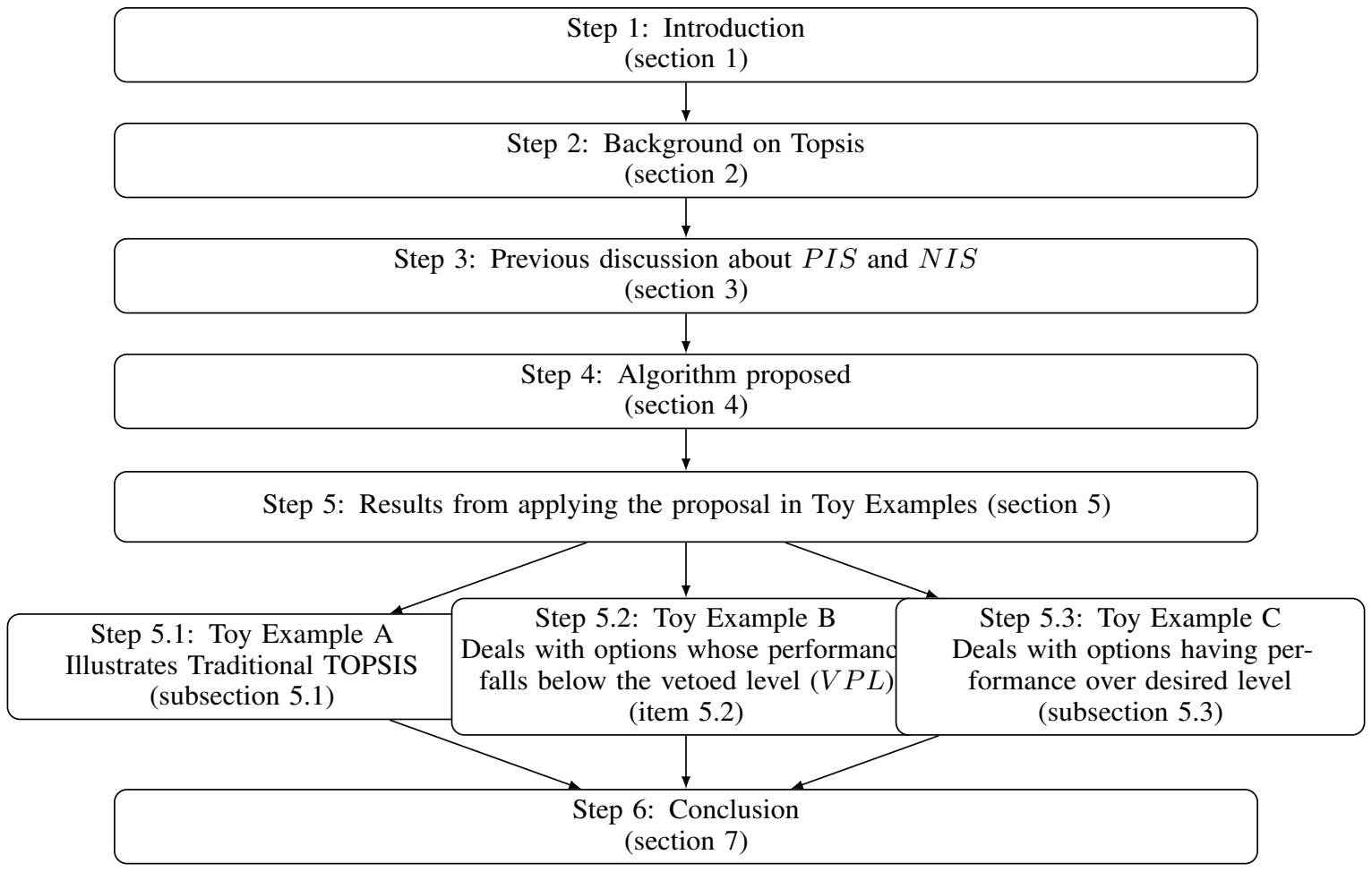

Figure 1: Graphical representation of the article structure

The remainder of the paper follows this sequence.

## 2 Background on TOPSIS seminal algorithm

TOPSIS, as described in Algorithm 1, was originally proposed by Hwang and Yoon [1981] as a multicriteria method for ranking alternatives according to their distances from a positive best frontier, or Positive Ideal Solution ($PIS$), and a worst frontier, or Negative Ideal Solution ($NIS$). Figure 2 displays the flowchart of TOPSIS Algorithm.

**Algorithm 1** TOPSIS, based on Hwang and Yoon [1981]

Step 1: **Get the inputs of the Decision Problem.** In this step the Decision Maker [DM] provides the inputs mentioned in Equation 1:

$$\begin{cases} A = \{a_1, a_2, ..., a_m\}; \\ C = \{c_1, c_2, ..., c_n\}; \\ G = \begin{pmatrix} g_{11} & g_{12} & \cdots & g_{1n} \\ g_{21} & g_{22} & \cdots & g_{2n} \\ \vdots & \vdots & \ddots & \vdots \\ g_{m1} & g_{m2} & \cdots & g_{mn} \end{pmatrix} \in \mathbb{R}^{m\times n}; \\ \mathbf{w} = (w_1, w_2, \ldots, w_n) \in \mathbb{R}^n, \qquad w_j \geq 0,\ \forall j \in \mathcal{J}, \qquad \sum_{j=1}^{n} w_j = 1 \\ \mathbf{p} = (p_1, p_2, \ldots, p_n) \in \{-1, 1\}^n \end{cases} \tag{1}$$

where:

- $A$ and $C$ are respectively the sets of $m$ alternatives set and $n$ criteria. Let $\mathcal{I} = \{1, \ldots, m\}$ and $\mathcal{J} = \{1, 2, \ldots, n\}$ be index sets.
- $G$ in the Decision Matrix, where $g_{ij}$ indicates de evaluation of alternative $a_i$ with respect to criterion $c_j$, for $i \in \mathcal{I}$ and $j \in \mathcal{J}$;
- $w_j$ is the weight of the criterion $c_j$, $\forall j \in \mathcal{J}$;
- $p_j = 1$ and $p_j = -1$ indicate respectively if criterion $c_j$ is beneficial or non-beneficial.

Step 2: **Normalize the criteria's weights**. In some situations, the imputed weights may be not normalized. In such situations it is necessary to normalize the weights ensuring that:

$$\sum_{j=1}^{j=n} w_j = 1 \tag{2}$$

Step 3: **Obtain the normalized matrix $R_{m\times n}$.** Apply a normalization procedure to the Decision Matrix to get $R \in \mathbb{R}^{m\times n}$. Among the normalization procedures, Hwang and Yoon [1981] used the one described in Equation 4.

$$R_{m\times n} = \begin{pmatrix} r_{11} & r_{12} & \cdots & r_{1n} \\ r_{21} & r_{22} & \cdots & r_{2n} \\ \vdots & \vdots & \ddots & \vdots \\ r_{m1} & r_{m2} & \cdots & r_{mn} \end{pmatrix} \tag{3}$$

Where:

$$r_{ij} = \frac{g_{ij}}{\sqrt{\sum_{i=1}^{m} g_{ij}^2}}\,, \quad \forall i \in \mathcal{I}, \forall j \in \mathcal{J}. \tag{4}$$

Step 4: **Weighting normalized matrix**. Compute the weighted normalized matrix $T \in {}^{\mathbb{R}}{}_{m\times n}$, by multiplying each column in $R$ by the the respective weight or constant of scale $w_j \in \mathbf{w}$.

$$T_{m\times n} = \begin{pmatrix} t_{11} & t_{12} & \cdots & t_{1n} \\ t_{21} & t_{22} & \cdots & t_{2n} \\ \vdots & \vdots & \ddots & \vdots \\ t_{m1} & t_{m2} & \cdots & t_{mn} \end{pmatrix} \tag{5}$$

Where:

$$t_{ij} = r_{ij} \times w_j\,, \quad \forall i \in \mathcal{I}, \forall j \in \mathcal{J} \tag{6}$$

Step 5: **Identify the normalized reference frontiers**. Get the reference arrays for the Positive Ideal Solution ($PIS$) and the negative Ideal Solution ($NIS$), given respectively by n-dimensional vectors, such that:

$$PIS = (t_1^+, ..., t_n^+), \text{ where } t_j^+ = \begin{cases} \max_{i \in \mathcal{I}} t_{ij}, & \text{if } p_j = 1 \\ \min_{i \in \mathcal{I}} t_{ij}, & \text{if } p_j = -1 \end{cases} \quad \forall j \in \mathcal{J} \tag{7}$$

$$NIS = (t_1^-, ..., t_n^-), \text{ where } t_j^- = \begin{cases} \min_{i \in \mathcal{I}} t_{ij}, & \text{if } p_j = 1 \\ \max_{i \in \mathcal{I}} t_{ij}, & \text{if } p_j = -1 \end{cases} \quad \forall j \in \mathcal{J} \tag{8}$$

Step 6: **Compute the distances from each alternative $a_i \in A$ to the positive and to negative ideal solution**

- Distance ($d_{ib}$) to the best or positive frontier $PIS$

$$d_{ib} = \sqrt{\sum_{j=1}^{n} (t_{ij} - PIS_j)^2}, \quad \forall i \in \mathcal{I} \tag{9}$$

- Distance ($d_{iw}$) to the worst or negative frontier $NIS$

$$d_{iw} = \sqrt{\sum_{j=1}^{n} (t_{ij} - NIS_j)^2}, \quad \forall i \in \mathcal{I} \tag{10}$$

Step 7: **Compute the relative closeness score $S_{iw}$ of each alternative** $a_i \in A$, that measures how much an alternative $a_i$ is close to the positive ideal solution, or, in other words: how much $a_i$ is away from the negative ideal solution?

$$S_{iw} = \frac{d_{ib}}{d_{ib} + +d_{iw}}, \quad \forall i \in \mathcal{I} \tag{11}$$

Step 8: **Rank alternatives**
Rank the alternatives according the rule: alternatives with higher values of $S_{iw}$ (computed by Equation 11), are put in the top of the ranking.

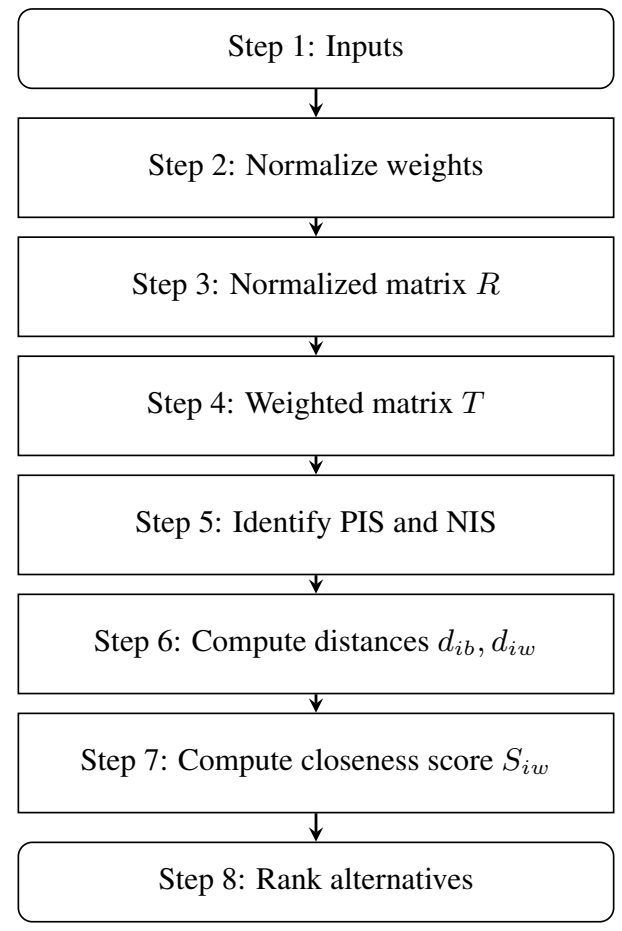

Figure 2: Flowchart of the TOPSIS algorithm steps, from input data to ranking of alternatives

As shown in Algorithm 1, the method reduces an $n$-dimensional evaluation problem to a bi-dimensional score function $S_{iw}$ (see Equation 11). Starting from the the alternatives performances $(g_{i1}, g_{i2}, \dots g_{in-1}, g_{in})$, where $g_{ij}$ denotes the performance of alternative $a_i$ on criterion $g_j$, TOPSIS ultimately works with two quantities only: $d_{ib}$ and $d_{iw}$, computed by Equation 10 and Equation 9, respectively.

In essence, TOPSIS rewards alternatives that remain close to the best frontier and distant from the worst one. That intuition is simple, but the way those frontiers are constructed is precisely what motivates the discussion in the next section.

## 3 Previous discussion about the definition of $PIS$ and $NIS$

This section reviews the main criticisms of TOPSIS discussed in section 1, focusing on how earlier studies treated the definition of $NIS$ and $PIS$ and on what those adjustments do, or do not, resolve.

Chen et al. [2011] argued that defining the Positive Ideal Solution ($PIS$) and the Negative Ideal Solution ($NIS$) directly from the observed dataset may generate undesirable effects, especially rank reversals when alternatives are added or removed. To mitigate that problem, the authors proposed fixing $PIS$ and $NIS$ as reference points in the consequence space, defined by the decision-maker and possibly informed by business benchmarks or known performance standards. They also departed from the usual TOPSIS workflow by applying normalization to the distances from those fixed reference points rather than to the raw decision matrix itself.

From a conceptual standpoint, Kong [2011] traced rank reversal in TOPSIS to the dependence of both normalization and ideal solutions on the set of available alternatives. The author therefore suggested fixing $PIS$ and $NIS$ exogenously, in line with the decision-maker's preferences. The paper also stressed that the normalization procedure should preserve independence among alternatives, showing that the original vector normalization used in TOPSIS does not satisfy that requirement and illustrating an alternative linear procedure based on max-min normalization with fixed $PIS$ and $NIS$ values.

García-Cascales and Lamata [2012a] reached a related conclusion. The authors showed that vector normalization strongly violates independence among alternatives and suggested linear normalization based on maximum criterion values as a partial remedy. Because that adjustment alone does not fully prevent rank reversal, they introduced an absolute mode of TOPSIS in which fictitious best and worst alternatives are added to stabilize $PIS$ and $NIS$.

Finally, de Farias Aires and Ferreira [2019] proposed a more systematic modification of the TOPSIS algorithm aimed at eliminating rank reversals by breaking the dependence among alternatives. Building on the preceding literature, the authors argued that $PIS$ and $NIS$ should be defined from a specified domain for each criterion and that normalization should depend exclusively on those domain limits. Under that view, normalized values become independent of the evaluated set itself. For example, if a benefit criterion is defined on the interval $[0, 10]$, then 0 and 10 determine, respectively, the $NIS$ and $PIS$, and a MaxMin or Max normalization uses only those limits to transform the corresponding column of the decision matrix.

Considering the above references, a reviewed version of TOPSIS would incorporate fixed PIS and NIS, and a normalization process that is not affected by changes in the alternative set.

---

**Algorithm 2** Reviewed TOPSIS based on Chen et al. [2011], Kong [2011], García-Cascales and Lamata [2012b], de Farias Aires and Ferreira [2019]

---

Step 1 → 2: These steps are similar to Step 1 and 2 from Algorithm 1.

Step 3: **Normalization and Weighting Process.** In this step, two modifications are done, regarding the Step 3 of Algorithm 1:

Step 3.a: Establish two additional performance vectors, which will be used both to obtain PIS and NIS later and in the normalization procedure of alternatives in $A$.

$$\begin{cases} \mathbf{g}^+ = (g_1^+, g_2^+, \dots, g_n^+) \in \mathbb{R}^n; \\ \mathbf{g}^- = (g_1^-, g_2^-, \dots, g_n^-) \in \mathbb{R}^n \end{cases} \tag{12}$$

where:

- $g_j^+$ and $g_j^-$ are respectively the values to the highest and lowest values of criterion $g_j$, and once they are defined, they could not change if the set of alternatives changes. In other words, they are reference or fixed values depends on the performance of the initial set of alternatives.

Step 3.b: Use a normalization process that is invariant to the inclusion or exclusion of alternatives in the original Decision Matrix. For that, use the values established in $\eth^+$ and $\eth^-$. For instance, one may use the following to obtain matrix $T \in \mathbb{R}^{m \times n}$:
Max-Min:

$$t_{ij} = \left( \frac{g_{ij} - g_j^-}{g_j^+ - g_j^-} \right) \times w_j \,, \quad \forall i \in \mathcal{I}, j \in \mathcal{J} \tag{13}$$

Steps 4: **PIS and NIS** Get PIS and NIS from the weighted normalized values of reference vectors $\eth^+$ and $\eth^-$, considering the directions defined in ı. For instance, weighted MaxMin normalization defined in Equation 13 maps these reference values to $w_j$ or zero. The resulting PIS and NIS, considering direction $p_j, \forall j \in \mathcal{J}$ will be:
PIS and NIS from Max-Min normalization:

$$PIS = (t_1^+, \ldots, t_n^+), \text{ where } t_j^+ = \begin{cases} w_j, & \text{if } p_j = 1 \\ 0, & \text{if } p_j = -1 \end{cases} \quad \forall j = 1, \ldots, n \tag{14}$$

$$NIS = (t_1^-, \ldots, t_n^-), \text{ where } t_j^- = \begin{cases} 0, & \text{if } p_j = 1 \\ w_j, & \text{if } p_j = -1 \end{cases} \quad \forall j = 1, \ldots, n \tag{15}$$

Steps 5 $\rightarrow$ 8: **Ranking Process.** Apply Steps 5, 6, 7, and 8 from Algorithm 1 to obtain the ranking of the alternatives based on their closeness coefficients.
We observe that although these innovations in TOPSIS provide important advance in avoiding ranking reversal, They have not protection against potential distortions that can be generated by eventual presence of outliers in the initial set of alternatives.

---

Therefore, taken together, these studies point to two central causes of rank reversal in TOPSIS: (i) the way $PIS$ and $NIS$ are defined, and (ii) the normalization process. They are especially relevant when criterion domains are known in advance and when the decision-maker can specify fixed and meaningful reference points. That assumption, however, is often demanding. In many decision problems, criterion domains are unbounded, poorly understood, or only partially known. In such cases, the arrival of a new alternative above $PIS$ or below $NIS$ raises an awkward modeling question: should the alternative be discarded, or should the domain and the ideal solutions be updated, thereby reopening the possibility of rank reversal? A related issue is that theoretical domains may stretch distances in a way that behaves like an outlier. Consider a grading system formally defined on $[0, 100]$ but with observed values between 15 and 40, where 35 is already regarded as a high level of performance. The concern is not confined to TOPSIS, however, and similar questions about ranking stability have also appeared in other MCDA families. Recent survey evidence indicates that rank reversal remains an active concern across several widely used MCDA methods and that methods prone to this phenomenon continue to be widely applied in practice [Li and Abbas, 2026]. Similar concerns have been examined outside TOPSIS, for example in WASPAS [Baykasoglu and Ercan, 2021] and AHP [Tu and Wu, 2025]. In VIKOR, the discussion has also reached the level of postprocessing reranking adjustments under additional fairness considerations [Dodevska et al., 2023].

Table 1 summarizes the main implications of this review. In short, the existing approaches reduce some forms of instability, but they do not fully address the noise introduced by extreme performances because normalization still depends, directly or indirectly, on relative magnitudes tied to the alternative set.

Table 1 identifies the main limitations left open by the previous proposals. To make the position of TOPSIS-RAD more explicit, Table 15 compares the present proposal with the main streams of work discussed above. The comparison also highlights how each approach deals, or fails to deal, with outlier influence.

The proposal developed in this paper takes a different route. Rather than focusing only on redefining the classical reference points, TOPSIS-RAD first screens alternatives with $VPL$, then caps performances with $DPL$, and only then ranks the transformed decision problem. In this sense, the contribution of the method is not limited to stabilizing the reference structure; it also changes which alternatives are allowed to influence the final ranking and how excessively high performances are treated.

Table 1: TOPSIS variants regarding the issues mentioned in section 1

| Issue | Feature | Source | Notes on gaps to be explored |
|---|---|---|---|
| Rank reversal | Solved by using external absolute reference levels based on external benchmarks. | Chen et al. [2011] | There remains a gap in addressing the impact of outliers, which can occur when an alternative exhibits performance substantially deviating from the established benchmark. |
| | Solved by allowing the DM to define the performance levels of the absolute reference points, while constraining these values to remain above/below the highest/lowest performances observed across the criteria set. | Kong [2011], García-Cascales and Lamata [2012a], Cables et al. [2016] | There is still a gap regarding outliers, which arise when an alternative exhibits performance significantly higher or lower than the rest of the options. |
| Let DM define the reference levels | Partially attended once the "freedom" to choose the absolute levels is limited by the performance of the alternatives in the set of options. | Kong [2011], García-Cascales and Lamata [2012a], Cables et al. [2016] | For example, it does not allow the DM to choose an absolute lower reference level that exceeds the lowest performance observed for any alternative within a given criterion. |

Table 2: Comparison between classical TOPSIS, previous TOPSIS variants, and TOPSIS-RAD

| Approach | DM-defined reference levels | Excludes infeasible alternatives before ranking | Caps performances above desired levels | Explicit mechanism to limit outlier influence | Main mechanism used against rank reversal / instability | Main remaining limitation |
|---|---|---|---|---|---|---|
| Classical TOPSIS (Hwang and Yoon [1981]) | No | No | No | No | None; ranking uses dataset-dependent $PIS$ and $NIS$ | Sensitive to rank reversal, outliers, and possible misalignment with DM requirements |
| Chen et al. [2011] | Yes, through fixed benchmark-like reference points | No | No | No | Replaces dataset-dependent ideal points with fixed references and normalizes distances to them | Does not explicitly prevent unacceptable alternatives or excessive performances from distorting the ranking |
| Kong [2011], García-Cascales and Lamata [2012a], Cables et al. [2016] | Partial; constrained by the observed set or by fictitious absolute references | No | No | Partial; indirect through stabilized references and normalization | Stabilizes ideal solutions and modifies normalization to reduce dependence on the alternative set | Still leaves limited freedom for the DM and no direct control at the ranking stage |
| de Farias Aires and Ferreira [2019] | Yes, through criterion-domain limits | No | No | Partial; indirect through fixed criterion domains | Breaks dependence among alternatives by defining $PIS$, $NIS$, and normalization from fixed criterion domains | Requires reliable domain knowledge and does not distinguish feasible from desired performance |
| TOPSIS-RAD | Yes, through $VPL$ and $DPL$ defined directly by the DM without reliance on bench- | Yes; $VPL$ removes non-viable alternatives *before* normalisation is computed | Yes; $DPL$ caps each performance at the DM's desired level, not at a theoretical domain bound | Yes; $VPL$ prevents non-viable outliers from entering normalisation; $DPL$ prevents desirable-but extreme | Anchors both the lower frontier (via $VPL$) and the upper frontier (via $DPL$) in explicit DM requirements, making the ranking independent of dataset extremes | Requires meaningful DM-specified reference levels; broader empirical validation and sensitivity analysis remain as future work |

## 4 Detailing the proposal

The novel TOPSIS-RAD is described in Algorithm 3. TOPSIS-RAD preserves the final distance-based logic of TOPSIS [Algorithm 1], but inserts and modifies some operations before that stage:

- Obtaining the Desired Performance Levels ($DPL$) and Vetoed Performance Levels ($VPL$) [Step 2;
- Determining the subset of qualified alternatives [Step 3];
- Capping the remaining alternatives by desired levels [Step 4];
- Applying a normalization procedure that preserves rank relations of remaining alternatives in cases where alternatives are included or deleted from the original set [Step 5];
- Combining this normalization with the $DPL$ and $VPL$ vectors to obtain Desired Performance Levels ($DNL$) and Vetoed Performance Levels ($VNL$). In other words, we define$DPL$ and $VPL$ as the numerical thresholds specified by the decision maker. After normalization, these thresholds are transformed into Desired Normalized Levels (DNL) and Vetoed Normalized Levels (VNL), which act as frontiers delimiting the feasible decision space. $DNL$ and $VPL$ are used as a substitute $PIS$ and $NIS$ vectors in such a way that these normalized frontiers are independent of the original alternative set $A$ or the Decision Matrix $G \in \mathbb{R}^{m\times n}$.

**Algorithm 3** TOPSIS-RAD Routine

Step 1: **Get the inputs of the Decision Problem.** This step is organized into two sub-steps, as it follows:

Step 1.a: In this step the Decision Maker [DM] provides the inputs mentioned in Equation 1, as described in Algorithm 1:

$$\begin{cases} A = \{a_1, a_2, ..., a_m\}; \\ C = \{c_1, c_2, ..., c_n\}; \\ G = \begin{pmatrix} g_{11} & g_{12} & \cdots & g_{1n} \\ g_{21} & g_{22} & \cdots & g_{2n} \\ \vdots & \vdots & \ddots & \vdots \\ g_{m1} & g_{m2} & \cdots & g_{mn} \end{pmatrix} \in \mathbb{R}^{m\times n}; \mathbf{w} = (w_1, w_2, \dots, w_n) \in \mathbb{R}^n, \qquad w_j \geq 0,\ \forall j \in \mathcal{J}, \qquad \sum_{j=1}^{n} w_j = 1 \\ \mathbf{p} = (p_1, p_2, \dots, p_n) \in \{-1, 1\}^n \end{cases}$$

where:

- $A$ and $C$ are respectively the sets of $m$ alternatives set and $n$ criteria. Let $\mathcal{I} = \{1, \dots, m\}$ and $\mathcal{J} = \{1, 2, \dots, n\}$ be index sets.
- $G$ in the Decision Matrix, where $g_{ij}$ indicates de evaluation of alternative $a_i$ with respect to criterion $c_j$, for $i \in \mathcal{I}$ and $j \in \mathcal{J}$;
- $w_j$ is the weight of the criterion $c_j$, $\forall j \in \mathcal{J}$;
- $p_j = 1$ and $p_j = -1$ indicate respectively if criterion $c_j$ is beneficial or non-beneficial.

Step 1.b: Obtain from the Decision-maker the following additional information:

- An array (DPL) with the performance levels that the DM wishes the selected alternatives to achieve.
- An array (VPL) that stores the minimum acceptable performance required for an alternative to remain in the feasible set.

$$\begin{cases} DPL = \{d_1, d_2, ..., d_n\}; \\ VPL = \{v_1, v_2, ..., v_n\}; \end{cases} \tag{16}$$

Step 2: **Normalize the weights.** As described Step 2 from Algorithm 1, it is necessary to normalize the weights ensuring that:

$$\sum_{j=1}^{j=n} w_j = 1$$

Step 3: **Obtain the normalized matrix $R_{k\times n}$.**

Step 3.a: Determine the set of qualified alternatives ($A^q$), by verifying the feasibility status of each alternative:

$$Feasible_i = \begin{cases} 1, & \text{if } v_{ij} = 0,\ \forall j = 1 \in \mathcal{J} \\ 0, & \text{otherwise} \end{cases} \qquad \forall i = i \in \mathcal{I} \tag{17}$$

The set $A^q$ of qualified alternatives will include alternatives $a_i \in A$ such that $Feasible_i = 1$).

$$A^q = \{a_i \in A : Feasible_i = 1\} \tag{18}$$

Step 3.b: Obtain a Qualified Decision Matrix ($G^q$) by retaining only the values of $a_k \in A^q$. Let $k = |A^q|$ be the number of qualified alternatives, $\mathcal{K} = \{i \in \mathcal{I} : a_i \in A^q\}$ the index set of qualified alternatives, and $\alpha(i)$ be a function that returns the i-th index in set $\mathcal{K}$. A restricted Qualified Decision Matrix is obtained:

$$G^q = \begin{pmatrix} g_{\alpha(1),1} & g_{\alpha(1),2} & \cdots & g_{\alpha(1),n} \\ g_{\alpha(2),1} & g_{\alpha(2),2} & \cdots & g_{\alpha(2),n} \\ \vdots & \vdots & \ddots & \vdots \\ g_{\alpha(k),1} & g_{\alpha(k),2} & \cdots & g_{\alpha(k),n} \end{pmatrix} \in \mathbb{R}^{k \times n} \tag{19}$$

Step 3.c: Update the Decision Matrix by clipping the performances in $G^q$ that are over DPL. Compare the values of matrix $G^q$ (i.e. performances of qualid fied alternatives) against the desired performance levels (DPL) to obtain the Desired-constrained matrix $G^d$:

$$g_{ij}^d = \begin{cases} \min(g_{ij}, d_j), & \text{if } p_j = 1 \\ \max(g_{ij}, d_j), & \text{if } p_j = -1 \end{cases} \qquad \forall\, i \in \mathcal{K},\ \forall j \in \mathcal{J} \tag{20}$$

$$G^d = \begin{pmatrix} g^d_{\alpha(1),1} & g^d_{\alpha(1),2} & \cdots & g^d_{\alpha(1),n} \\ g^d_{\alpha(2),1} & g^d_{\alpha(2),2} & \cdots & g^d_{\alpha(2),n} \\ \vdots & \vdots & \ddots & \vdots \\ g^d_{\alpha(k),1} & g^d_{\alpha(k),2} & \cdots & g^d_{\alpha(k),n} \end{pmatrix} \in \mathbb{R}^{k \times n} \tag{21}$$

Step 3.d: Compute the normalized matrix $R$ using the unaccepted (or vetoed) and desired levels as lower and upper bounds, respectively:

$$r_{ij} = \frac{g_{ij}^d - v_j}{d_j - v_j} \qquad \forall i \in \mathcal{K}, j \in \mathcal{J} \tag{22}$$

$$R = \begin{pmatrix} r_{\alpha(1),1} & r_{\alpha(1),2} & \cdots & r_{\alpha(1),n} \\ r_{\alpha(2),1} & r_{\alpha(2),2} & \cdots & r_{\alpha(2),n} \\ \vdots & \vdots & \ddots & \vdots \\ r_{\alpha(k),1} & r_{\alpha(k),2} & \cdots & r_{\alpha(k),n} \end{pmatrix} \in \mathbb{R}^{k \times n} \tag{23}$$

Step 4: Compute the weighted normalized matrix $T$:

$$t_{ij} = w_j r_{ij} \qquad \forall i \in \mathcal{K}, j \in \mathcal{J} \tag{24}$$

$$T = \begin{pmatrix} t_{\alpha(1),1} & t_{\alpha(1),2} & \cdots & t_{\alpha(1),n} \\ t_{\alpha(2),1} & t_{\alpha(2),2} & \cdots & t_{\alpha(2),n} \\ \vdots & \vdots & \ddots & \vdots \\ t_{\alpha(k),1} & t_{\alpha(k),2} & \cdots & t_{\alpha(k),n} \end{pmatrix} \in \mathbb{R}^{k \times n} \tag{25}$$

Step 5: Define the normalized solutions: Desired Normalized Level (DNL) and Veto Normalized Level (VNL), which are obtained directly from the weighted normalized desired and vetoed reference vectors. Since the normalization in Step 5 maps the desired level $d_j$ and the unaccepted level $v_j$ to 1 or 0 depending on the specific criterion direction, the weighted normalized reference values are $w_j$ and 0, respectively. Therefore, the DLN and VNL are given by:

$$DNL = \mathbf{t}^+ = (t_1^+, t_2^+, \ldots, t_n^+), \quad \text{where } t_j^+ = \begin{cases} w_j, & \text{if } p_j = 1 \\ 0, & \text{if } p_j = -1 \end{cases} \qquad \forall j \in \mathcal{J} \tag{26}$$

$$VNL = \mathbf{t}^- = (t_1^-, t_2^-, \ldots, t_n^-) \quad \text{where } t_j^- = \begin{cases} 0, & \text{if } p_j = 1 \\ w_j, & \text{if } p_j = -1 \end{cases} \qquad \forall j \in \mathcal{J} \tag{27}$$

Step 6: Compute the Euclidean distances of each retained alternative to the $D^+$ and $U^-$:

$$d_{ib} = \sqrt{\sum_{j=1}^{n}(t_{ij} - t_j^+)^2} \qquad \forall i \in \mathcal{K} \tag{28}$$

$$d_{iw} = \sqrt{\sum_{j=1}^{n}(t_{ij} - t_j^-)^2} \qquad \forall i \in \mathcal{K} \tag{29}$$

Step 7: Compute the closeness coefficient ($S_{iw}$) of each retained alternative:

$$S_{iw} = \frac{d_{iw}}{d_{ib} + d_{iw}} \qquad \forall i \in \mathcal{K} \tag{30}$$

Step 8: Rank the retained alternatives in decreasing order of $Sw_i$.

Unlike PIS and NIS in TOPSIS, which define positive and negative ideal solutions based on the extremes of the dataset, we propose metrics grounded in normative levels: the Desired Normalized Level (DLN) and the Veto Normalized Level (VNL). Alternatives falling below the VNL are automatically excluded, reflecting an absolute veto criterion.

## 5 Results from applying the proposal

This section uses toy examples to compare the behavior of traditional TOPSIS with that of TOPSIS-RAD. The examples are not meant to provide broad empirical validation; their role is narrower, namely, to show how the introduction of Desired Performance Levels ($DPL$) and Vetoed Performance Levels ($VPL$) changes the construction of the ranking problem described in section 4. Toy Example A establishes the baseline using classical TOPSIS and introduces the dataset on which all three examples are computed. Toy Example B addresses the second paper objective (Objective 2 in section 1) by applying $VPL$ to exclude a non-viable alternative and demonstrating how dataset-driven reference points change when a violating alternative is removed. Toy Example C addresses the third paper objective (Objective 3 in section 1) by applying $DPL$ to cap performances above the desired level, stabilising the normalisation frontiers and limiting the influence of extreme performances on the final ranking.

For simplicity and without loss of genality, the numerical examples consider equal weights ($w_j = \frac{1}{n}, \forall j \in \mathcal{J}$) and only benefit criteria ($p_j = 1, \forall j \in \mathcal{J}$) and are given equal weights. These choices keep the comparison centered on the methodological differences between traditional TOPSIS and the proposed approach, rather than on weighting issues or criterion-direction effects. However, in practical applications, the definition of criteria weights plays an important role in reflecting the decision-maker's preferences. The literature provides a wide range of structured approaches for eliciting such weights. For instance, Rezaei [2015, 2016] proposed the Best–Worst Method (BWM), which determines weights through the solution of an optimization model based on pairwise comparisons between each criterion and the most and least important criteria, requiring fewer comparisons than the traditional AHP [Saaty et al., 1980] while ensuring a high level of consistency. Other approaches are also widely adopted, such as the SMARTER method [Edwards and Barron, 1994] and the Rank Order Centroid (ROC) weights [Barron, 1992, Barron and Barrett, 1996].

The toy examples were implemented by using the Visual TOPSIS-RAD, a web app that we developed to support decision modeling based on both: traditional TOPSIS and TOPSIS-RAD algorithms.The Figure 3 shows the first screen of such app, which we made tool available for free access at `https://topsis-ranking.vercel.app/`.

### 5.1 Applying traditional TOPSIS (Toy example A)

In this subsection, all steps of traditional TOPSIS [Hwang and Yoon, 1981] are applied to an example named here as Toy example A.

Step 1: **Get the initial data.**

1. Alternatives: $A = \{A_1, A_2, A_3, A_4, A_5, A_6, A_7, A_8, A_9, A_{10}\}$
2. Criteria: $C = \{c_1, c_2, c_3, c_4\}$, corresponding to *performance* ($c_1$), *quality* ($c_2$), *sustainability* ($c_3$), and *maturity* ($c_4$)
3. $m = 10, n = 4, \mathcal{I} = \{1, 2, \dots, 10\}, \mathcal{J} = \{1, 2, 3, 4\}$
4. Weights: $\mathbf{w} = (0.25, 0.25, 0.25, 0.25)$
5. Preference drivers: $\mathbf{p} = (1, 1, 1, 1)$, all criteria are "benefit".

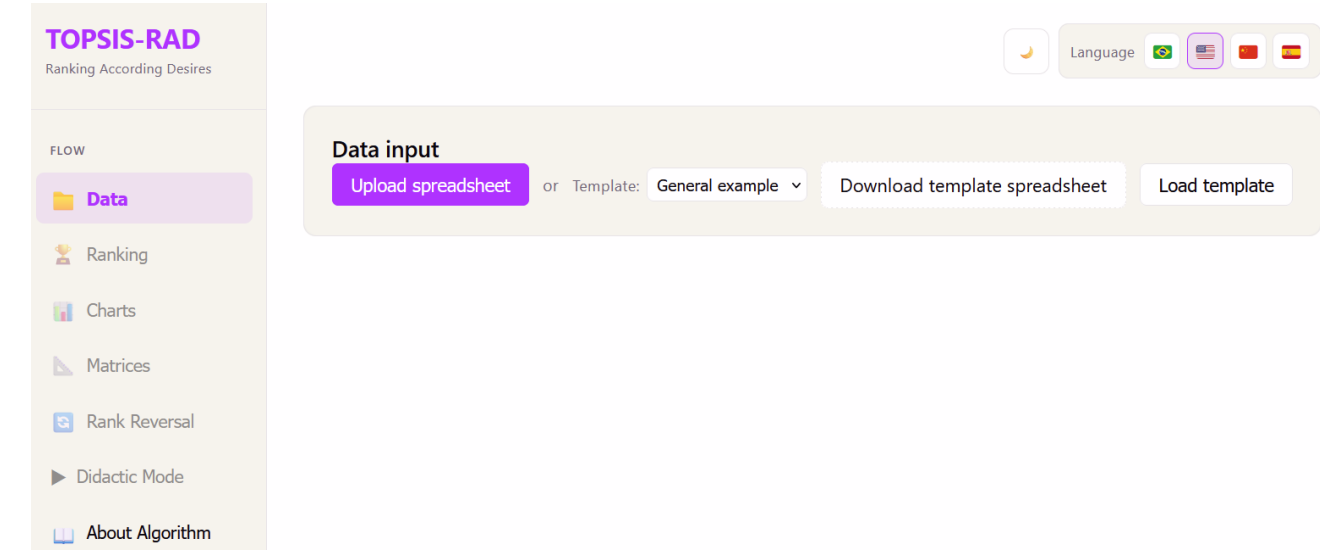

Figure 3: First screen of Visual TOPSIS-RAD

6. Pay-off matrix ($G \in \mathbb{R}^{m \times n}$):
Table 3 shows the pay-off matrix $G$. The values of $A$, $C$ and $G$ were read from a csv file (`dataset_Tosis_Rad_Csv.csv`) uploaded into Visual TOPSIS-RAD. The other inputs ($W$, $P$) are the default values loaded in the app – these can be changed by the user at any time. Figure 4 anticipates the final ranking produced by Visual TOPSIS-RAD for Toy Example A; the step-by-step derivation follows in Steps 2–8.

Table 3: Pay-off matrix $G$ for Toy Example A ($m = 10$ alternatives, $n = 4$ criteria).

| | $C_1$ | $C_2$ | $C_3$ | $C_4$ |
|---|---|---|---|---|
| $A_1$ | 68 | 72 | 74 | 78 |
| $A_2$ | 62 | 76 | 78 | 80 |
| $A_3$ | 58 | 70 | 150 | 76 |
| $A_4$ | 80 | 72 | 99 | 90 |
| $A_5$ | 90 | 88 | 47 | 118 |
| $A_6$ | 80 | 64 | 66 | 70 |
| $A_7$ | 76 | 68 | 98 | 105 |
| $A_8$ | 80 | 50 | 5 | 41 |
| $A_9$ | 70 | 74 | 78 | 75 |
| $A_{10}$ | 60 | 68 | 82 | 81 |

Step 2: **Obtain the normalized weights** $W$**.** As each criterion $w_j \in W$ had the same weight, it resulted in $w_j = 0.25$ for all $j \in \mathcal{J}$.

Step 3: **Obtain the normalized matrix** $R_{m \times n}$**.** Among the available normalization procedures, Visual TOPSIS-RAD applies the min-max range normalization described in Equation 22. Table 4 displays the values of $r_{ij}$ computed from the data of Step 1.

Step 4: **Compute the weighted normalized matrix** $T_{m \times n}$**.** As all criteria have equal weight, this results in the values also shown in Table 4.

Table 4: Normalised matrix $R$ and weighted matrix $T$ for Toy Example A.

| | $R$ (normalised) | | | | $T = 0.25 \cdot R$ | | | |
|---|---|---|---|---|---|---|---|---|
| | $C_1$ | $C_2$ | $C_3$ | $C_4$ | $C_1$ | $C_2$ | $C_3$ | $C_4$ |
| A1 | 0.3125 | 0.5789 | 0.4759 | 0.4805 | 0.0781 | 0.1447 | 0.1190 | 0.1201 |
| A2 | 0.1250 | 0.6842 | 0.5034 | 0.5065 | 0.0313 | 0.1711 | 0.1259 | 0.1266 |
| A3 | 0.0000 | 0.5263 | 1.0000 | 0.4545 | 0.0000 | 0.1316 | 0.2500 | 0.1136 |
| A4 | 0.6875 | 0.5789 | 0.6483 | 0.6364 | 0.1719 | 0.1447 | 0.1621 | 0.1591 |
| A5 | 1.0000 | 1.0000 | 0.2897 | 1.0000 | 0.2500 | 0.2500 | 0.0724 | 0.2500 |
| A6 | 0.6875 | 0.3684 | 0.4207 | 0.3766 | 0.1719 | 0.0921 | 0.1052 | 0.0942 |
| A7 | 0.5625 | 0.4737 | 0.6414 | 0.8312 | 0.1406 | 0.1184 | 0.1603 | 0.2078 |
| A8 | 0.6875 | 0.0000 | 0.0000 | 0.0000 | 0.1719 | 0.0000 | 0.0000 | 0.0000 |
| A9 | 0.3750 | 0.6316 | 0.5034 | 0.4416 | 0.0938 | 0.1579 | 0.1259 | 0.1104 |
| A10 | 0.0625 | 0.4737 | 0.5310 | 0.5195 | 0.0156 | 0.1184 | 0.1328 | 0.1299 |

Step 5: **Identify the normalized arrays $PIS$ and $NIS$.** As $p_j = 1 \ \forall j \in \mathcal{J}$, we have a positive direction. The application of Equations (7) and (8) yields the values shown in Table 5.

Table 5: $PIS$ and $NIS$ arrays for Toy Example A.

| | $C_1$ | $C_2$ | $C_3$ | $C_4$ |
|---|---|---|---|---|
| $PIS$ | 0.2500 | 0.2500 | 0.2500 | 0.2500 |
| $NIS$ | 0.0000 | 0.0000 | 0.0000 | 0.0000 |

Step 6: **Compute the distances of each alternative $a_i \in A$ to the positive ($PIS$) and negative ($NIS$) ideal solutions.** The results obtained by applying Equation 9 and Equation 10 are displayed in Table 6.

Step 7: **Compute the relative closeness score ($S_{iw}$) of each alternative $a_i \in A$ with respect to the negative ideal solution $NIS$.** The results obtained by applying Equation 11 are shown in Table 6.

Step 8: **Rank the alternatives.** According to the original TOPSIS procedure (Algorithm 1) proposed by Hwang and Yoon [1981], alternatives are ranked using the values of the $S_{iw}$ score. Table 6 shows the ranking positions of the complete set of alternatives in $A$ while Figure 4 illustrates the scores obtained by the alternatives.

Table 6: Distances to $PIS$/$NIS$, scores $S_{iw}$, and ranking for Toy Example A.

| | $d_{ib}$ (to $PIS$) | $d_{iw}$ (to $NIS$) | $S_{iw}$ | Rank |
|---|---|---|---|---|
| $A_1$ | 0.2732 | 0.2359 | 0.4633 | 7 |
| $A_2$ | 0.2911 | 0.2492 | 0.4613 | 8 |
| $A_3$ | 0.3084 | 0.3045 | 0.4968 | 4 |
| $A_4$ | 0.1822 | 0.3195 | 0.6369 | 2 |
| $A_5$ | 0.1776 | 0.4390 | 0.7120 | 1 |
| $A_6$ | 0.2762 | 0.2407 | 0.4657 | 6 |
| $A_7$ | 0.1977 | 0.3204 | 0.6184 | 3 |
| $A_8$ | 0.4400 | 0.1719 | 0.2809 | 10 |
| $A_9$ | 0.2604 | 0.2485 | 0.4883 | 5 |
| $A_{10}$ | 0.3169 | 0.2208 | 0.4107 | 9 |

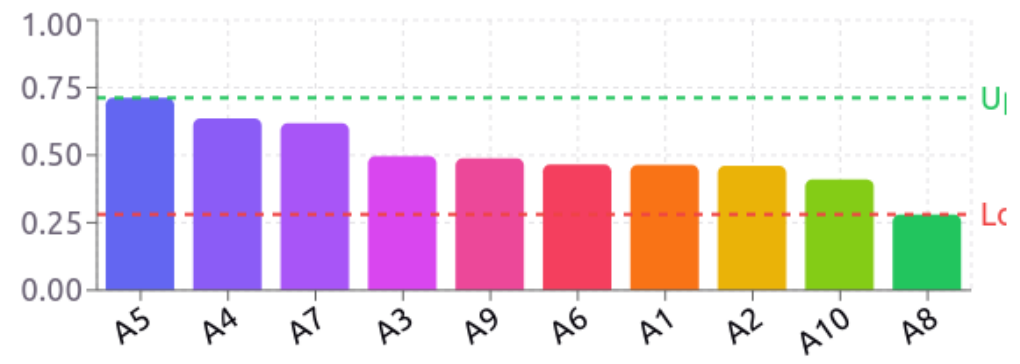

Figure 4: Final ranking by $S_{iw}$ score (Toy Example A).

### 5.2 Applying TOPSIS-RAD with VPL (Toy Example B)

In this Toy Example B we apply the TOPSIS-RAD described in Algorithm 3 to the same dataset used in Toy Example A (see Table 3). The purpose is to isolate one specific effect: how an alternative that fails the Vetoed Performance Level ($VPL$) threshold can, through its extreme values, distort the ranking of the alternatives that remain eligible. The $DPL$ is set equal to the best observed performances (column maxima of the full dataset), and the $VPL$ is designed so that exactly one alternative is excluded.

Step 1: **Get the inputs of the Decision Problem.**

Step 1.a: Get the initial data. The same performance matrix $G$ used in Toy Example A is adopted here; it appears in Table 3. The dataset contains ten alternatives ($A_1$–$A_{10}$) evaluated on four benefit criteria ($C_1$–$C_4$) with equal weights $w_j = 0.25$.

Step 1.b: Get *Desired Performance Levels* ($DPL$) and *Vetoed Performance Levels* ($VPL$) from the Decision Maker.
The $DPL$ and $VPL$ values are shown in Table 7. The $DPL$ is set equal to the maximum value of each column of the Decision Matrix while the $VPL$ is set at: $VPL = \{20,\ 20,\ 20,\ 20\}$. Because $A_8$ presents performances below the $VPL$ in criteria $c_2$, it fails the veto check and is excluded from the qualified set.

Table 7: TOPSIS-RAD settings for Toy Example B: $VPL$ (lower bound) and $DPL$ (upper bound) per criterion.

| | $C_1$ | $C_2$ | $C_3$ | $C_4$ |
|---|---|---|---|---|
| $VPL$ | 20 | 20 | 20 | 20 |
| $DPL$ | 90 | 88 | 150 | 118 |

Step 2: **Obtain the normalised weights $W$.** To allow comparison with Toy Example A, the same equal weights are used: $w_j = 0.25$ for all $j \in \mathcal{J}$.

Step 3: **Obtain the qualified and desired-constrained normalized decision matrix.**

Step 3.a: **Determine the *qualified* alternatives ($A^q$).** Comparing each alternative's performance against the $VPL$ vector reveals that $A_8$ violates the veto on criterion $C_3$ ($g_{8,2} = 5 < 20$). All remaining alternatives satisfy the thresholds, so:

$$A^q = \{A_1,\ A_2,\ A_3,\ A_4,\ A_5,\ A_6,\ A_7,\ A_9,\ A_{10}\}.$$

Step 3.b: **Obtain the Qualified Decision Matrix ($G^q$) by retaining only the rows of $a_k \in A^q$.** $G^q$ is obtained from Table 3 by removing the row of $A_8$, yielding a $9 \times 4$ matrix.

Step 3.c: **Clip performances above $DPL$ to obtain $G^d$.** Since all values in $G^q$ are at most equal to the corresponding $DPL$ entry (the column maximum of the original dataset), no clipping is required in this example and $G^d = G^q$.

Step 3.d: **Compute the normalized matrix $R_{k\times n}$ .** Applying the normalization formula $r_{ij} = (g^d_{ij} - VPL_j)/(DPL_j - VPL_j)$ of Equation 22 to $G^d_{k\times n}$ it resulted in the values that appear in the left side of Table 8.

Step 4: **Compute the weighted normalized matrix $T_{k\times n}$.** Multiplying the values computed in the previous step by the weights $w_j = 0.25$ yields the values of $T$ shown in the right side of Table 8.

Table 8: Normalised matrix $R$ and weighted matrix $T$ for Toy Example B.

| | $R$ (normalised) | | | | $T = 0.25 \cdot R$ | | | |
|---|---|---|---|---|---|---|---|---|
| | $C_1$ | $C_2$ | $C_3$ | $C_4$ | $C_1$ | $C_2$ | $C_3$ | $C_4$ |
| $A_1$ | 0.6857 | 0.7647 | 0.4154 | 0.5918 | 0.1714 | 0.1912 | 0.1038 | 0.1480 |
| $A_2$ | 0.6000 | 0.8235 | 0.4462 | 0.6122 | 0.1500 | 0.2059 | 0.1115 | 0.1531 |
| $A_3$ | 0.5429 | 0.7353 | 1.0000 | 0.5714 | 0.1357 | 0.1838 | 0.2500 | 0.1429 |
| $A_4$ | 0.8571 | 0.7647 | 0.6077 | 0.7143 | 0.2143 | 0.1912 | 0.1519 | 0.1786 |
| $A_5$ | 1.0000 | 1.0000 | 0.2077 | 1.0000 | 0.2500 | 0.2500 | 0.0519 | 0.2500 |
| $A_6$ | 0.8571 | 0.6471 | 0.3538 | 0.5102 | 0.2143 | 0.1618 | 0.0885 | 0.1276 |
| $A_7$ | 0.8000 | 0.7059 | 0.6000 | 0.8673 | 0.2000 | 0.1765 | 0.1500 | 0.2168 |
| $A_9$ | 0.7143 | 0.7941 | 0.4462 | 0.5612 | 0.1786 | 0.1985 | 0.1115 | 0.1403 |
| $A_{10}$ | 0.5714 | 0.7059 | 0.4769 | 0.6224 | 0.1429 | 0.1765 | 0.1192 | 0.1556 |

Step 5: **Identify the Desired Normalised Level ($DNL$) and Vetoed Normalised Level ($VNL$) arrays.** The $DNL$ and $VNL$ are obtained directly from the weighted normalized $DPL$ and $VPL$. Considering that all criteria have a positive direction, Equations (26) and (27) will result in:

$$DNL = \{0.25,\ 0.25,\ 0.25,\ 0.25\}, \qquad VNL = \{0.0,\ 0.0,\ 0.0,\ 0.0\}.$$

Step 6: **Compute the distances of each qualified alternative $a_k \in A^q$ to the $DNL$ and $VNL$ reference arrays.** Using the Euclidean distance formulas of Algorithm 1, the values of $d_{ib}$ and $d_{iw}$ shown in Table 9 are obtained.

Step 7: **Compute the relative closeness score ($S_{iw}$) and rank the alternatives.** From the distances of Step 6, the score $S_{iw} = d_{iw}/(d_{iw} + d_{ib})$ is calculated for each alternative. The resulting values are shown in Table 9; the bar-chart ranking is shown in Figure 5.

Table 10 places the two rankings side by side, making the effect of the $VPL$ filter immediately visible. It reveals that using fixed $VPL$/$DPL$ frontiers produces a different ranking from traditional TOPSIS.

Table 9: Distances to $DNL$/$VNL$, scores $S_{iw}$, and ranking for Toy Example B.

| | $d_{ib}$ (to $DNL$) | $d_{iw}$ (to $VNL$) | $S_{iw}$ | Rank |
|---|---|---|---|---|
| $A1$ | 0.2035 | 0.3140 | 0.6068 | 7 |
| $A2$ | 0.2013 | 0.3174 | 0.6119 | 6 |
| $A3$ | 0.1701 | 0.3676 | 0.6837 | 4 |
| $A4$ | 0.1395 | 0.3707 | 0.7266 | 2 |
| $A5$ | 0.1981 | 0.4361 | 0.6877 | 3 |
| $A6$ | 0.2239 | 0.3101 | 0.5807 | 9 |
| $A7$ | 0.1379 | 0.3750 | 0.7312 | 1 |
| $A9$ | 0.1974 | 0.3216 | 0.6197 | 5 |
| $A_{10}$ | 0.2071 | 0.3000 | 0.5916 | 8 |

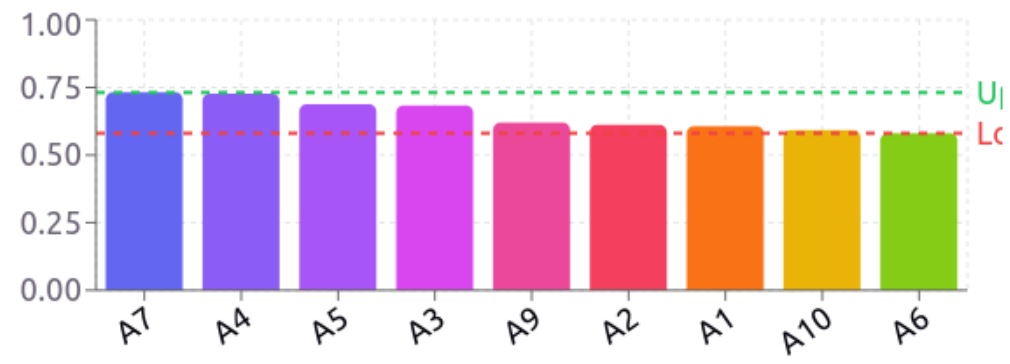

Figure 5: Final ranking by $S_{iw}$ score for Toy Example B: $A_7$ rises to 1st place; fixed $VPL$/$DPL$ frontiers anchor normalisation boundaries.

Alternative $A_7$, which ranked third under traditional TOPSIS, rises to first place once the $VPL$ is introduced into the system. Note that the effect of introducing $VPL = \{20, 20, 20, 20\}$ goes beyond removing alternative $A_8$. It introduces a frontier that, besides possibly vetoing some alternative (such as $A_8$), works as a reference point for the calculation of the normalized matrix and for obtaining the $VNL$. For instance, instead of using 60 as the "Min" reference point for criterion $C_1$ in the normalization, now $VPL_1 = 20$ is used. The same four alternatives (A3, A4, A5, A7) occupy the top positions in both examples, though in different order.

Table 10: Comparing rankings: traditional TOPSIS (Toy Example A) versus TOPSIS-RAD with $VPL$ (Toy Example B).

| | Ranking position | | | | | | | | | |
|---|---|---|---|---|---|---|---|---|---|---|
| Toy Example | $1^{st}$ | $2^{nd}$ | $3^{rd}$ | $4^{th}$ | $5^{th}$ | $6^{th}$ | $7^{th}$ | $8^{th}$ | $9^{th}$ | $10^{th}$ |
| A (TOPSIS) | A5 | A4 | A7 | A3 | A9 | A6 | A1 | A2 | A10 | A8 |
| B (VPL) | A7 | A4 | A5 | A3 | A9 | A2 | A1 | A10 | A6 | — |

A key property of fixed $VPL$/$DPL$ frontiers is that they anchor the normalisation boundaries at DM-specified levels rather than at dataset extremes. Because the normalisation formula $r_{ij} = (g^d_{ij} - VPL_j)/(DPL_j - VPL_j)$ uses the fixed $VPL$ and $DPL$ values—rather than the column extremes of the observed alternatives—removing $A_8$ does not shift the normalisation range of any criterion. The scores of all remaining alternatives are computed against the same fixed scale before and after the veto, even though the ranking changes to reflect the removal of the outlier alternative. Therefore, inserting or removing an alternative from the Decision Matrix does not cause ranking reversals. Observe that it is precisely a property that data-driven normalisation cannot guarantee. The effect of anchoring the upper frontier at a fixed $DPL$ is further illustrated in Toy Example C.

### 5.3 Applying TOPSIS-RAD with fixed DPL (Toy Example C)

In Toy Example C we apply TOPSIS-RAD to the same dataset used in Toy Examples A and B, with a $VPL$ set equal to the column minima of the full dataset and a *fixed*, decision-maker-defined $DPL$. The purpose is to show that anchoring the normalisation upper frontier to an explicit saturation threshold— below the dataset maximum—equalises over-performing alternatives and shifts the ranking towards well-balanced profiles.

Step 1: **Get the inputs of the Decision Problem.**

Step 1.a: Get the initial data. The same performance matrix and weights used in Toy Example A are used here.

Step 1.b: Get *Desired Performance Levels* ($DPL$) and *Vetoed Performance Levels* ($VPL$) from the Decision Maker.
The $VPL$ is now set equal to the column minima of the full dataset. Therefore, no alternative is excluded by the veto threshold ($g_{i,j} \geq VPL_j$ for all $i$ and $j$). The $DPL$ is set by the decision maker as fixed saturation thresholds: $DPL = \{80,\ 80,\ 80,\ 80\}$. Crucially, note that $DPL_j = 80$ is below the observed maximum on every criterion, so several alternatives will be clipped in the next step. The settings are shown in Table 11.

Table 11: TOPSIS-RAD settings for Toy Example C: $VPL$ (lower bound) and fixed $DPL$ (upper bound) per criterion.

| | $C_1$ | $C_2$ | $C_3$ | $C_4$ |
|---|---|---|---|---|
| $VPL$ | 58 | 50 | 5 | 41 |
| $DPL$ | 80 | 80 | 80 | 80 |

Step 2: **Obtain the normalised weights.** The same equal weights are used: $w_j = 0.25$ for all $j \in \mathcal{J}$. As the weights already sum 1, this step does not change the weights vector.

Step 3: **Obtain the qualified and desired-constrained decision matrix.**

Step 3.a: **Determine the *qualified* alternatives ($A^q$).** All the alternatives satisfy the $VPL$ thresholds, so no alternative is excluded:

$$A^q = A = \{A_1,\ A_2,\ A_3,\ A_4,\ A_5,\ A_6,\ A_7,\ A_8,\ A_9,\ A_{10}\}.$$

Step 3.b: **Obtain the Qualified Decision Matrix ($G^q$)** by retaining all ten rows (the full matrix from Table 3).

Step 3.c: **Clip performances above $DPL$ to obtain $G^d$.** With $DPL_j = 80$ for all criteria, any value exceeding 80 is capped. The affected entries are: $g^d_{3,3} = \min(150, 80) = 80$, $g^d_{4,3} = \min(99, 80) = 80$, $g^d_{4,4} = \min(90, 80) = 80$, $g^d_{5,1} = \min(90, 80) = 80$, $g^d_{5,2} = \min(88, 80) = 80$, $g^d_{5,4} = \min(118, 80) = 80$, $g^d_{7,3} = \min(98, 80) = 80$, $g^d_{7,4} = \min(105, 80) = 80$, $g^d_{10,3} = \min(82, 80) = 80$, $g^d_{10,4} = \min(81, 80) = 80$. All other values lie at or below 80 and are unchanged.

Step 3.d: **Compute the normalised matrix $R_{k\times n}$ .** Applying the normalisation formula $r_{ij} = (g^d_{ij} - VPL_j)/(DPL_j - VPL_j)$ of Equation 22, the obtained normalized matrix values are shown in the left side of Table 12.

Step 4: **Compute the the weighted normalised matrix** By multiplying the values in $G^d$ by $w_j = 0.25$, matrix $T$ is obtained, as shown inTable 12. The ten alternatives are included in this example. The key changes relative to Toy Example A are the capping of $A_5$, $A_3$, $A_4$, $A_7$, and $A_{10}$ on multiple criteria, which equalises their top scores, as detailed in **Step 3.c**.

Table 12: Normalised matrix $R$ and weighted matrix $T$ for Toy Example C.

| | $R$ (normalised) | | | | $T = 0.25 \cdot R$ | | | |
|---|---|---|---|---|---|---|---|---|
| | $C_1$ | $C_2$ | $C_3$ | $C_4$ | $C_1$ | $C_2$ | $C_3$ | $C_4$ |
| $A_1$ | 0.4545 | 0.7333 | 0.9200 | 0.9487 | 0.1136 | 0.1833 | 0.2300 | 0.2372 |
| $A_2$ | 0.1818 | 0.8667 | 0.9733 | 1.0000 | 0.0455 | 0.2167 | 0.2433 | 0.2500 |
| $A_3$ | 0.0000 | 0.6667 | 1.0000 | 0.8974 | 0.0000 | 0.1667 | 0.2500 | 0.2244 |
| $A_4$ | 1.0000 | 0.7333 | 1.0000 | 1.0000 | 0.2500 | 0.1833 | 0.2500 | 0.2500 |
| $A_5$ | 1.0000 | 1.0000 | 0.5600 | 1.0000 | 0.2500 | 0.2500 | 0.1400 | 0.2500 |
| $A_6$ | 1.0000 | 0.4667 | 0.8133 | 0.7436 | 0.2500 | 0.1167 | 0.2033 | 0.1859 |
| $A_7$ | 0.8182 | 0.6000 | 1.0000 | 1.0000 | 0.2045 | 0.1500 | 0.2500 | 0.2500 |
| $A_8$ | 1.0000 | 0.0000 | 0.0000 | 0.0000 | 0.2500 | 0.0000 | 0.0000 | 0.0000 |
| $A_9$ | 0.5455 | 0.8000 | 0.9733 | 0.8718 | 0.1364 | 0.2000 | 0.2433 | 0.2179 |
| $A_{10}$ | 0.0909 | 0.6000 | 1.0000 | 1.0000 | 0.0227 | 0.1500 | 0.2500 | 0.2500 |

Step 5: **Identify the Desired Normalised Level ($DNL$) and Vetoed Normalised Level ($VNL$) arrays.** The fixed $DPL$/$VPL$ frontiers guarantee that the normalised scale spans exactly $[0, 1]$ on each criterion. The reference arrays are:

$$DNL = \{0.25,\ 0.25,\ 0.25,\ 0.25\}, \qquad VNL = \{0,\ 0,\ 0,\ 0\}.$$

Step 6: **Compute the distances of each qualified alternative $a_k \in A^q$ to the $DNL$ and $VNL$ arrays.** The Euclidean distances are given in Table 13.

Step 7: **Compute the relative closeness score ($S_{iw}$) and rank the alternatives.** The resulting scores and final ranking are shown in Table 13; the bar-chart ranking is shown in Figure 6.

Table 13: Distances to $DNL$/$VNL$, scores $S_{iw}$, and ranking for Toy Example C.

| | $d_{ib}$ (to $DNL$) | $d_{iw}$ (to $VPL$) | $S_{iw}$ | Rank |
|---|---|---|---|---|
| $A_1$ | 0.1536 | 0.3946 | 0.7197 | 5 |
| $A_2$ | 0.2074 | 0.4132 | 0.6659 | 7 |
| $A_3$ | 0.2648 | 0.3750 | 0.5861 | 9 |
| $A_4$ | 0.0667 | 0.4702 | 0.8758 | 1 |
| $A_5$ | 0.1100 | 0.4551 | 0.8053 | 2 |
| $A_6$ | 0.1551 | 0.3899 | 0.7154 | 6 |
| $A_7$ | 0.1098 | 0.4351 | 0.7984 | 3 |
| $A_8$ | 0.4330 | 0.2500 | 0.3660 | 10 |
| $A_9$ | 0.1284 | 0.4066 | 0.7600 | 4 |
| $A_{10}$ | 0.2483 | 0.3847 | 0.6078 | 8 |

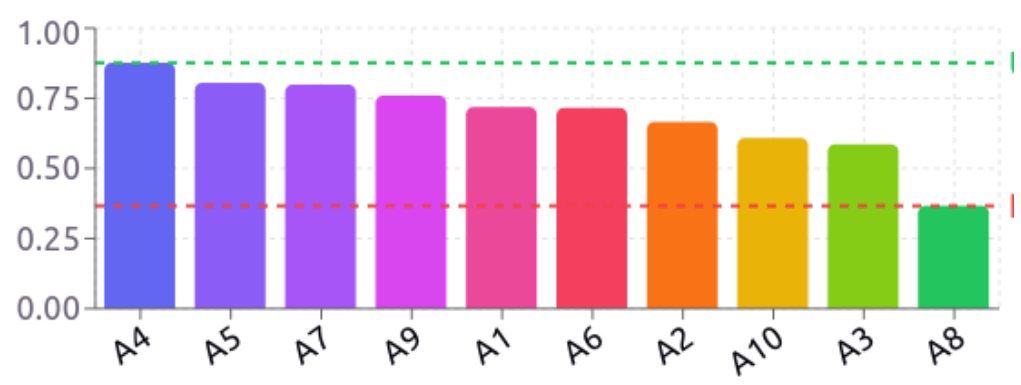

Figure 6: Final ranking by $S_{iw}$ score for Toy Example C: $A_4$ rises to 1st place. $A_8$ is not excluded, and occupies the to 10th.

Table 14 places the three rankings side by side. It shows that setting $DPL = 80$ on all criteria causes a dramatic reshaping of the ranking. Alternative $A_4 = \{80, 72, 99, 90\}$ rises to first place with the highest $S_{iw}$ score of 0.8758. Note that this alternative presents performances above $VPL$ in criteria $C_3$ and $C_4$ (both clipped to 80). In addition, this alternative originally scores 80 in criterion $C_1$. Alternative $A_7$ moves to the third place, as it loses the advantage of obtaining a score of 105 in criterion $C_4$, 25 units above $DPL_4 = 80$. The driving mechanism is that $DPL = 80$ eliminates every above-threshold advantage simultaneously. Any alternative whose strength was concentrated in values above 80 on a few criteria is equalised at the cap and can no longer compensate for weaknesses in other criteria. Conversely, alternatives with a profile that is uniformly close to 80 across all criteria, such as $A_4$, gain substantially.

Table 14: Comparing rankings: traditional TOPSIS (Toy Example A), TOPSIS-RAD with data-driven $DPL$ (Toy Example B), and TOPSIS-RAD with fixed $DPL$ (Toy Example C).

| | Ranking position | | | | | | | | | |
|---|---|---|---|---|---|---|---|---|---|---|
| Toy Example | $1^{st}$ | $2^{nd}$ | $3^{rd}$ | $4^{th}$ | $5^{th}$ | $6^{th}$ | $7^{th}$ | $8^{th}$ | $9^{th}$ | $10^{th}$ |
| A (TOPSIS) | A5 | A4 | A7 | A3 | A9 | A6 | A1 | A2 | A10 | A8 |
| B (VPL) | A7 | A4 | A5 | A3 | A9 | A2 | A1 | A10 | A6 | — |
| C (RAD, fixed DPL$=80$) | A4 | A5 | A7 | A9 | A1 | A6 | A2 | A10 | A3 | A8 |

The critical advantage of fixed $DPL$ frontiers is *invariance to dataset composition*. In Toy Example B, the upper normalisation frontier on each criterion was anchored at the fixed $DPL$ (column maxima), so removing $A_8$ did not shift the normalisation range of any criterion—the frontiers remained at the DM-specified levels. In Toy Example C the same property holds: any alternative with a performance above 80 is simply capped; any alternative with a performance at or below 80 is normalised against the fixed span $[VPL_j, 80]$. Adding or removing alternatives from the evaluated set does not shift any frontier, and the ranking therefore reflects only the performances of the evaluated alternatives against the fixed scale—precisely the property that data-driven normalisation cannot guarantee.

To illustrate this point, suppose a new alternative $A_{11} = \{25, 30, 25, 45\}$ is included into the original data set. Table 15 reports the closeness coefficients $S_{iw}$ and the resulting rankings under four configurations: traditional TOPSIS and the three TOPSIS-RAD specifications combining the $DPL$ and $VPL$ used in Toy Examples B and C. Each computed twice: once for the original set $A = \{A_1, \ldots, A_{10}\}$ and once for the augmented set $A \cup \{A_{11}\}$. Comparing the two columns within each method isolates the effect that inserting $A_{11}$ has on the evaluation of the pre-existing alternatives. Under traditional TOPSIS, this single insertion alters every closeness coefficient and reshuffles the ranking: $A_7$ rises from third to first, $A_5$ falls from first to second, and $A_6$ drops from sixth to ninth, even though $A_{11}$ itself is ranked last. This is a clear instance of rank reversal, as an alternative dominated by all others changes their relative order merely by entering the dataset. It happens because the $PIS$, the $NIS$, and the normalisation that depends on them are all recomputed from the new column extremes. By contrast, under every TOPSIS-RAD specification the closeness coefficients of the original alternatives are identical before and after the inclusion of $A_{11}$, so their ranking is preserved exactly. The new alternative is either appended at the bottom of the ranking when it satisfies the veto thresholds (last place under $VPL = \{20, 20, 20, 20\}$) or excluded altogether when it does not (vetoed under $VPL = \{58, 50, 5, 41\}$, since its performances on $C_1$ and $C_2$ fall below the required minima). This invariance is a direct consequence of the fixed $VPL/DPL$ frontiers: because the normalisation span $[VPL_j, DPL_j]$ does not depend on the alternative set, adding or removing an alternative cannot shift any criterion's scale and therefore cannot reverse the order of the surviving alternatives.

Table 15: Effect of inserting alternative $A_{11}$ on TOPSIS and TOPSIS-RAD rankings

| | TOPSIS | | | | TOPSIS-RAD | | | | TOPSIS-RAD | | | | TOPSIS-RAD | | | |
|---|---|---|---|---|---|---|---|---|---|---|---|---|---|---|---|---|
| | | | | | $VPL = \{20, 20, 20, 20\}$ $DPL = \{90, 88, 150, 118\}$ | | | | $VPL = \{58, 50, 5, 41\}$ $DPL = \{80, 80, 80, 80\}$ | | | | $VPL = \{20, 20, 20, 20\}$ $DPL = \{80, 80, 80, 80\}$ | | | |
| **Alt.** | $A$ | | $A \cup \{A_{11}\}$ | | $A$ | | $A \cup \{A_{11}\}$ | | $A$ | | $A \cup \{A_{11}\}$ | | $A$ | | $A \cup \{A_{11}\}$ | |
| | $S_{iw}$ | Rank | $S_{iw}$ | Rank | $S_{iw}$ | Rank | $S_{iw}$ | Rank | $S_{iw}$ | Rank | $S_{iw}$ | Rank | $S_{iw}$ | Rank | $S_{iw}$ | Rank |
| $A_1$ | 0.4633 | 7 | 0.5815 | 7 | 0.6068 | 7 | 0.6068 | 7 | 0.7197 | 5 | 0.7197 | 5 | 0.8709 | 4 | 0.8709 | 4 |
| $A_2$ | 0.4613 | 8 | 0.5880 | 6 | 0.6119 | 6 | 0.6119 | 6 | 0.6659 | 7 | 0.6659 | 7 | 0.8545 | 5 | 0.8545 | 5 |
| $A_3$ | 0.4968 | 4 | 0.6359 | 4 | 0.6837 | 4 | 0.6837 | 4 | 0.5861 | 9 | 0.5861 | 9 | 0.8084 | 8 | 0.8084 | 8 |
| $A_4$ | 0.6369 | 2 | 0.7067 | 3 | 0.7266 | 2 | 0.7266 | 2 | 0.8758 | 1 | 0.8758 | 1 | 0.9356 | 1 | 0.9356 | 1 |
| $A_5$ | 0.7120 | 1 | 0.7120 | 2 | 0.6877 | 3 | 0.6877 | 3 | 0.8053 | 2 | 0.8053 | 2 | 0.7649 | 9 | 0.7649 | 9 |
| $A_6$ | 0.4657 | 6 | 0.5505 | 9 | 0.5807 | 9 | 0.5807 | 9 | 0.7154 | 6 | 0.7154 | 6 | 0.8109 | 7 | 0.8109 | 7 |
| $A_7$ | 0.6184 | 3 | 0.7207 | 1 | 0.7312 | 1 | 0.7312 | 1 | 0.7984 | 3 | 0.7984 | 3 | 0.8989 | 2 | 0.8989 | 2 |
| $A_8$ | 0.2809 | 10 | 0.3685 | 10 | vetoed | | vetoed | | 0.3660 | 10 | 0.3660 | 10 | vetoed | | vetoed | |
| $A_9$ | 0.4883 | 5 | 0.5924 | 5 | 0.6197 | 5 | 0.6197 | 5 | 0.7600 | 4 | 0.7600 | 4 | 0.8943 | 3 | 0.8943 | 3 |
| $A_{10}$ | 0.4107 | 9 | 0.5603 | 8 | 0.5916 | 8 | 0.5916 | 8 | 0.6078 | 8 | 0.6078 | 8 | 0.8188 | 6 | 0.8188 | 6 |
| $A_{11}$ | – | – | 0.0717 | 11 | – | – | 0.1485 | 10 | – | – | vetoed | | – | – | 0.2197 | 10 |

It should be emphasised that the traditional TOPSIS results in Table 15 were obtained using the MaxMin normalisation rather than the vector normalisation originally prescribed in Algorithm 1. Holding the normalisation procedure fixed across all configurations ensures that any difference between traditional TOPSIS and TOPSIS-RAD stems from the use of $DPL$ and $VPL$ and not from the normalisation scheme. MaxMin normalisation alone, however, does not prevent rank reversal. As the traditional TOPSIS column shows, inserting $A_{11}$ still reshuffles the ranking, just as it could happen under the original vector normalisation (noted effect in the TOPSIS literature, as explored in Section 3). Reversals are avoided only when the MaxMin normalisation is "locked" to fixed reference values, such as when we use $DPL$ and $VPL$, so that its bounds no longer depend on the alternative set, and those same references are additionally used to define the ideal ($DNL$) and anti-ideal ($VNL$) solutions. It is therefore the combination of fixed normalisation bounds and reference-based ideal/anti-ideal points, not the MaxMin formula in isolation, that grants TOPSIS-RAD its immunity to rank reversal.

Finally, Table 15 also illustrates a modelling capability that classical TOPSIS lacks: in TOPSIS-RAD it is the decision maker's aspiration and veto levels, rather than the dataset, that determine which performance profiles are rewarded. Alternative $A_5 = \{90, 88, 47, 118\}$ makes this concrete. For the two configurations where $DPL = \{80, 80, 80, 80\}$, this alternative ranks second under $VPL = \{58, 50, 5, 41\}$ yet falls to last among the qualified alternatives under $VPL = \{20, 20, 20, 20\}$. Because both configurations apply the same $DPL$ cap, this shift stems entirely from the $VPL$. The $DPL = 80$ cap first removes the raw magnitude advantage of $A_5$: three of its performances exceed 80 (on $C_1$, $C_2$, and $C_4$) and are clipped to the cap, leaving only its weak $C_3 = 47$ to carry information within the $[VPL_j, 80]$ span. Lowering the floor to 20 then widens the normalisation span on the capped criteria to $[20, 80]$, compressing almost every alternative into the upper portion of those scales. Reaching the cap ceases to be distinctive, the three strengths of $A_5$ no longer separate it from the field, and its isolated $C_3$ weakness is exposed and penalised by the Euclidean distance, which squares each criterion-wise gap. The opposite happens to $A_4 = \{80, 72, 99, 90\}$, whose only sub-cap performance ($C_2 = 72$) is high: once the cap neutralises every above-threshold advantage, its balanced profile prevails and it rises to first place. This is intended behaviour, not instability. By saturating performances at the desired level, the $DPL$ enacts a satisficing logic in which exceeding an aspiration earns no further credit; by removing alternatives that fall below the minimum acceptable level, the $VPL$ imposes a veto mechanism. Together they allow the decision maker to favour alternatives that meet the desired level across all criteria over those whose appeal rests on a few performances far above it—a preference that classical TOPSIS cannot represent, and one that distinguishes TOPSIS-RAD from earlier fixed-reference variants, which stabilise the ranking frontiers but neither screen out non-viable alternatives nor saturate performance at an explicitly desired level.

# 6 Discussion

## 6.1 When TOPSIS-RAD is most beneficial

TOPSIS-RAD is designed for decision contexts in which the DM can articulate, prior to the evaluation, both a minimum acceptable level of performance and a desired (but not necessarily maximum) level of performance for each criterion. In such contexts, using dataset-driven $PIS$ and $NIS$ has two well-documented drawbacks. First, the presence of even a single alternative with an extreme value can dominate the normalisation frontiers, compressing the scores of all other alternatives on that criterion—effectively hiding useful performance differences. Second, the rankings depend implicitly on which alternatives happen to have been included in the evaluation, a property that undermines stability when the set is revised.

Both concerns are directly addressed when meaningful $VPL$ and $DPL$ values are available. The three toy examples in section 5 illustrate the mechanism: once fixed $VPL$/$DPL$ frontiers are applied, removing $A_8$ via the veto filter does not shift the normalisation range of any criterion, even though the ranking changes to reflect the removal of the outlier (Toy Example B). Conversely, when a fixed $DPL$ cap is set below the observed column maxima, alternatives whose strength lies in performances above the cap are equalised, and the ranking shifts towards well-balanced profiles (Toy Example C).

TOPSIS-RAD is therefore particularly suited to recurring evaluations (supplier selection, project portfolios, personnel assessments) where the DM's performance standards are relatively stable, and to regulated contexts (e.g., procurement or certification) where minimum thresholds are formally defined.

## 6.2 Sensitivity to $VPL$ and $DPL$ specification

A key practical question is how sensitive the TOPSIS-RAD ranking is to the choice of reference levels. Two types of sensitivity deserve attention.

First, $VPL$ *sensitivity*: tightening a $VPL$ threshold removes more alternatives and may substantially reshape the normalisation frontiers for the surviving set. Decision makers should consider running the evaluation under alternative $VPL$ settings to assess how robust the ranking is to their minimum requirements. In the extreme case where every alternative satisfies all $VPL$ thresholds, TOPSIS-RAD reduces to a standard TOPSIS calculation with $DPL$-capped performances.

Second, $DPL$ *sensitivity*: if $DPL$ values are set well above the observed column maxima, the capping has no effect and the result again converges to standard TOPSIS. If $DPL$ values are set below the column maxima (as in Toy Example C), several alternatives are equalised at the $DPL$ frontier, and performance differences below that frontier become the sole basis for discrimination. Sensitivity analysis over $DPL$ is therefore equivalent to asking: "how much does the ranking depend on performances beyond the desired level?" The answer is directly visible in the normalised matrix $R$.

Both types of sensitivity analysis can be conducted interactively using the Visual TOPSIS-RAD web application described in section 5.

## 6.3 Relationship to reference-point methods in MCDA

TOPSIS-RAD belongs to a broader class of methods that anchor the evaluation in DM-defined reference points rather than in dataset extremes. The use of aspiration and reservation levels is a central theme in several established MCDA frameworks. In ELECTRE TRI [Yu, 1992], for example, alternatives are assigned to categories defined by fixed reference profiles rather than ranked relative to each other, which provides full independence from the alternative set. In NADIR–UTOPIA formulations within multiobjective optimisation, the DM specifies an ideal and an anti-ideal point that define the normalisation space [Miettinen, 1999]. The satisficing concept in behavioural decision theory [Simon, 1956] similarly distinguishes acceptable from unacceptable outcomes through explicit thresholds.

TOPSIS-RAD can be seen as bridging the TOPSIS tradition and these reference-point approaches: it retains the distance-based aggregation and the familiar $S_{iw}$ score of TOPSIS while replacing the data-driven frontiers with DM-specified boundaries. Unlike ELECTRE TRI, it does not assign alternatives to predefined categories but produces a complete cardinal ranking of the surviving alternatives. Unlike NADIR–UTOPIA normalisation, it explicitly removes non-viable alternatives before the scoring stage rather than projecting them onto a fixed scale.

This positioning may help practitioners who are already familiar with TOPSIS to adopt a more preference-driven workflow without abandoning the computational structure they know.

## 7 Conclusion

This study introduced TOPSIS-RAD as a TOPSIS variant in which the decision maker specifies two reference arrays before the final ranking stage. The method uses $VPL$ to remove non-viable alternatives and $DPL$ to cap performances above the desired level before normalisation and distance-based ranking. This change directly targets three recurring difficulties in traditional TOPSIS: weak alignment with DM requirements, sensitivity to extreme values, and rank instability driven by dataset-dependent reference points.

The results in section 5 clarify where this difference comes from. In Toy Example A, traditional TOPSIS placed $A_5$ in first place, with $A_4$ second and $A_7$ third. In Toy Example B, once $VPL$ filtered out $A_8$, which violated the vetoed threshold on criterion $C_3$ ($g_{8,3} = 5 < VPL_3 = 20$), the ranking changed to place $A_7$ in first, with $A_4$ second, $A_5$ third, and $A_3$ fourth. The same four alternatives occupy the top positions, though in different order. The point is that fixed $VPL$/$DPL$ frontiers anchor the normalisation boundaries at DM-specified levels rather than at dataset extremes, so the normalisation range does not shift when a non-viable alternative is removed.

Toy Example C highlights a different mechanism. With a uniform fixed $DPL = 80$ across all criteria, the performances of $A_5$ (clipped on $C_1$, $C_2$, and $C_4$), $A_3$, $A_4$, $A_7$, and $A_{10}$ (clipped on $C_3$ and/or $C_4$) are capped before normalisation, equalising the top performers on those criteria. As a result, $A_4$, which has a uniformly balanced profile close to the $DPL$, rises from second to first place with the highest $S_{iw}$ score (0.8758), displacing $A_5$, whose first-place standing in Toy Example A rested on three values well above 80 (on $C_1$, $C_2$, and $C_4$) that are now equalised at the cap. The capping effect is even more pronounced for $A_3$: its fourth-place standing in Toy Example A was driven by an exceptional performance of 150 on $C_3$ (almost twice the desired level). So once this value is capped at 80, $A_3$ loses its main advantage and falls to ninth place. The point is not to penalise strong performance, but to prevent values far above what is practically desired from dominating the entire ranking.

The two components of TOPSIS-RAD address these issues in different ways:

- **Vetoed Performance Levels ($VPL$)** — Enable the exclusion of non-viable alternatives that fail to meet minimum performance thresholds, preventing them from distorting the reference points and introducing noise into the rankings of qualified alternatives.
- **Desired Performance Levels ($DPL$)** — Define upper bounds that reflect the DM's aspirations rather than dataset extremes, preventing outliers from disproportionately influencing the ranking process while preserving the consideration of high-performing alternatives.

Unlike traditional TOPSIS, where the ranking is entirely driven by dataset extremes, TOPSIS-RAD introduces decision-maker-defined reference levels before the final distance-based scoring stage. The resulting evaluation remains recognizable as TOPSIS, but it is anchored in reference levels that reflect the actual decision context and the DM's expectations.

This distinction matters most in decision contexts where:

- Minimum performance standards must be enforced (e.g., regulatory compliance, safety requirements, quality assurance);
- Extreme performances introduce noise rather than meaningful information (e.g., resource allocation with budget constraints, personnel selection with realistic job requirements);
- Decision-maker preferences and aspirations should guide the evaluation rather than dataset characteristics (e.g., strategic planning, policy selection).

### 7.1 Limitations and Future Research

While TOPSIS-RAD improves robustness and incorporates DM preferences more directly than traditional TOPSIS, it also introduces additional complexity. The method requires the DM to specify $VPL$ and $DPL$ values explicitly, which demands deeper engagement with the decision problem and may increase cognitive effort, particularly in high-dimensional settings. It also assumes that the DM has enough domain knowledge to define meaningful reference levels, an assumption that may be demanding in highly uncertain or rapidly evolving environments.

The most immediate extensions of TOPSIS-RAD concern the following points:

(a) **Integration with uncertainty modeling** — Extend the framework to handle fuzzy, interval-valued, or probabilistic inputs for both criteria performances and reference levels, addressing decision contexts characterized by imprecision or incomplete information.

(b) **Sensitivity analysis and robustness testing** — Investigate the stability of TOPSIS-RAD rankings under variations in $VPL$, $DPL$, and criterion weights.

(c) **Comparative empirical studies** — Conduct extensive comparisons with other robust MCDM methods across diverse real-world applications.

(d) **Group decision-making extensions** — Adapt the method to accommodate multiple DMs with potentially divergent reference levels and preferences, exploring strategies for dealing with collective $VPL$ and $DPL$ specification.

## ACKNOWLEDGEMENTS

**Conflict of Interest**

The authors declare no conflicts of interest.

**Data Svailability Statement**

The data that support the findings are attached as supplementary files.

**Funding**

This study was financed in part by the Coordenação de Aperfeiçoamento de Pessoal de Nível Superior (CAPES), Finance Code 001; by the Conselho Nacional de Desenvolvimento Científico e Tecnológico (CNPq), Grant No. 314737/2026-0; and by the Fundação Carlos Chagas Filho de Amparo à Pesquisa do Estado do Rio de Janeiro (FAPERJ), Grant No. E-26 200.408/2026.

**Use of Generative AI**

The authors used generative AI tools (ChatGPT and Claude) to assist with orthographical, grammatical editing, and minor corrections to LaTeX scripts. These tools were used only for writing and coding support. All conceptual, methodological, analytical, and interpretive decisions, as well as the final review and approval of the manuscript, remained the responsibility of the authors.

## References

Ching-Lai Hwang and Kwangsun Yoon. *Multiple Attribute Decision Making: Methods and Applications*. Springer-Verlag, 1981.

Majid Behzadian, S. Khanmohammadi Otaghsara, Morteza Yazdani, and Joshua Ignatius. A state-of the-art survey of topsis applications. *Expert Systems with Applications*, 39(17):13051–13069, 2012. doi:10.1016/j.eswa.2012.05.056.

Vitor Anes and António Abreu. Adaptive cluster-based normalization for robust topsis in multicriteria decision-making. *Multicriteria Decision-Making. Applied Sciences*, 15(7), 2025. doi:10.3390/app15074044.

Feng Kong. Rank reversal and rank preservation in topsis. In H Zhang, G Shen, and D Jin, editors, *ADVANCED RESEARCH ON INDUSTRY, INFORMATION SYSTEMS AND MATERIAL ENGINEERING, PTS 1-7*, volume 204-210 of *Advanced Materials Research*, pages 36–41. Int Sci & Educ Res Assoc; Beijing Spon Res Inst; Beijing Gireida Educ Co Ltd, 2011. ISBN 978-3-03785-027-5. doi:10.4028/www.scientific.net/AMR.204-210.36. International Conference on Industry, Information System and Material Engineering, Guangzhou, PEOPLES R CHINA, APR 16-17, 2011.

Ye Chen, D. Marc Kilgour, and Keith W. Hipel. An extreme-distance approach to multiple criteria ranking. *Mathematical and Computer Modelling*, 53:646–658, 3 2011. ISSN 08957177. doi:10.1016/j.mcm.2010.10.001.

M. Socorro García-Cascales and M. Teresa Lamata. On rank reversal and topsis method. *Mathematical and Computer Modelling*, 56(5):123–132, 2012a. ISSN 0895-7177. doi:https://doi.org/10.1016/j.mcm.2011.12.022.

Renan Felinto de Farias Aires and Luciano Ferreira. A new approach to avoid rank reversal cases in the topsis method. *COMPUTERS & INDUSTRIAL ENGINEERING*, 132:84–97, JUN 2019. ISSN 0360-8352. doi:10.1016/j.cie.2019.04.023.

M. Socorro García-Cascales and M. Teresa Lamata. On rank reversal and topsis method. *Mathematical and Computer Modelling*, 56(5-6):123–132, 2012b. doi:10.1016/j.mcm.2011.12.022.

Bitong Li and Ali Abbas. A trend analysis of rank reversal in widely used decision-making methods. *Journal of Multi-Criteria Decision Analysis*, 33(1):e70027, 2026. doi:10.1002/mcda.70027.

Adil Baykasoglu and Elif Ercan. Analysis of rank reversal problems in "weighted aggregated sum product assessment" method. *Soft Computing*, 25(24):15243–15254, 2021. doi:10.1007/s00500-021-06405-w.

Jiancheng Tu and Zhibin Wu. Analytic hierarchy process rank reversals: causes and solutions. *Annals of Operations Research*, 346(2):1785–1809, 2025. doi:10.1007/s10479-023-05278-6.

Zorica Dodevska, Andrija Petrovic, Sandro Radovanovic, and Boris Delibasic. Changing criteria weights to achieve fair vikor ranking: a postprocessing reranking approach. *Autonomous Agents and Multi-Agent Systems*, 37(1):9, 2023. doi:10.1007/s10458-022-09591-5.

E. Cables, M. T. Lamata, and J. L. Verdegay. Rim-reference ideal method in multicriteria decision making. *INFORMATION SCIENCES*, 337:1–10, APR 10 2016. ISSN 0020-0255. doi:10.1016/j.ins.2015.12.011.

Jafar Rezaei. Best-worst multi-criteria decision-making method. *Omega*, 53:49–57, 2015.

Jafar Rezaei. Best-worst multi-criteria decision-making method: Some properties and a linear model. *Omega*, 64:126–130, 2016.

Thomas L Saaty et al. *The analytic hierarchy process*. McGraw-Hill, New York, 1980.

Ward Edwards and F Hutton Barron. Smarts and smarter: Improved simple methods for multiattribute utility measurement. *Organizational behavior and human decision processes*, 60(3):306–325, 1994.

F Hutton Barron. Selecting a best multiattribute alternative with partial information about attribute weights. *Acta psychologica*, 80(1-3):91–103, 1992.

F Hutton Barron and Bruce E Barrett. Decision quality using ranked attribute weights. *Management science*, 42(11):1515–1523, 1996.

W. Yu. ELECTRE TRI: Aspects méthodologiques et manuel d'utilisation. Technical report, Document du LAMSADE No. 74, Université Paris-Dauphine, 1992.

Kaisa Miettinen. *Nonlinear Multiobjective Optimization*. Kluwer Academic Publishers, Boston, 1999.

Herbert A. Simon. Rational choice and the structure of the environment. *Psychological Review*, 63(2): 129–138, 1956.